%% file: main.tex
\pdfoutput=1

\documentclass[11pt,a4paper]{article}
\usepackage[hyperref]{eacl2021}
\usepackage{breakurl}
\usepackage{times}
\usepackage{latexsym}
\usepackage{longtable}

\usepackage{microtype}

\aclfinalcopy % Uncomment this line for the final submission
\title{Analysing The Impact Of Linguistic Features On Cross-Lingual Transfer}

\author{Błażej Dolicki \quad Gerasimos Spanakis \\
  Maastricht University \\ Maastricht, The Netherlands \\
  \texttt{blazej.dol@gmail.com} \\
  \texttt{jerry.spanakis@maastrichtuniversity.nl} \\}

\date{}

\begin{document}
\maketitle
\begin{abstract}There is an increasing amount of evidence that in cases with little or no data in a target language, training on a different language can yield surprisingly good results. However, currently there are no established guidelines for choosing the training (source) language. In attempt to solve this issue we thoroughly analyze a state-of-the-art multilingual model and try to determine what impacts good transfer between languages. As opposed to the majority of multilingual NLP literature, we don't only train on English, but on a group of almost 30 languages. We show that looking at particular syntactic features is 2-4 times more helpful in predicting the performance than an aggregated syntactic similarity. We find out that the importance of syntactic features strongly differs depending on the downstream task - no single feature is a good performance predictor for all NLP tasks. As a result, one should not expect that for a target language $L_1$ there is a single language $L_2$ that is the best choice for any NLP task (for instance, for Bulgarian, the best source language is French on POS tagging, Russian on NER and Thai on NLI). We discuss the most important linguistic features affecting the transfer quality using statistical and machine learning methods.
\end{abstract}

\section{Introduction}
A vast majority of currently available NLP datasets is in English.  However, in many applications around the world it is crucial to use the local language. At the same time, creating new datasets requires a lot of resources and time. This is why recently multilingual language models are gaining more and more attention in academia and industry. Using them allows to train well-performing models with little or no data in the target language (usually low-resourced) given a corresponding dataset in the source language (usually high-resourced). This technique is called zero-shot learning.

We explore how differences in linguistic features correlate with or potentially affect performance on various kinds of downstream tasks. To present how such analysis can be helpful, let us consider the following scenario: We want to train a model in a task $T$ in language $L_1$, but we do not have access to a dataset for $T$ in $L_1$. However, we have n datasets for $T$ in languages $L_2$, $L_3$, ... $L_{n+1}$. We want to choose one of these languages that has the best `transferability' to $L_1$ for this specific task $T$. Currently, such a decision is based on intuition (either linguistic or researcher experience). This paper aims to provide an empirical framework on how to tackle such a scenario.

Over the past years several models for zero-shot learning were introduced such as LASER \cite{LASER},  MultiFit \cite{Multifit} or XLM-R \cite{xlmr}. For our analysis, we decided to use the XLM-R model as it achieves the best performance. To facilitate our experiments we take advantage of XTREME \cite{xtreme} which  is a recently published benchmark for multilingual models covering 40 languages on 8 diverse NLP tasks. However, it allows only training on English. To boost research in multilingual direction, we extend this benchmark to make it suitable for training in languages other than English and make the code publicly available\footnote{https://github.com/blazejdolicki/multilingual-analysis}. Additionally, we publish results of XLM-R Base model for 3 out of 8 tasks, for all available languages, to encourage further analysis of the nature of interactions between languages. 

To conclude, this work aims to examine the connection between linguistic features and the quality of transfer between languages on a variety of downstream tasks.

The following research questions are posed:
\begin{itemize}
\item Does the inclusion of fine-grained linguistic features improve prediction of language transfer compared to an aggregated syntactic distance between languages?
\item Which linguistic features have the largest impact on `transferability' and performance on 3 different tasks?
\item Can we identify some general linguistic features that determine good transfer for many tasks or is the feature importance task-specific?
\end{itemize}

\section{Related work}
\subsection{Zero-shot learning}
Multilingual language models during training create a common vector space for all languages used. This makes them especially relevant for zero-shot learning. The goal is to achieve similar vectors for sentences or words in different languages which are semantically equivalent (i.e. which are translations of each other). 

One of the first multilingual models that gained traction in academic circles was m-BERT \cite{mBERT}. It provided a shared vocabulary for all languages and was trained on a concatenation of corpora in about 100 different languages. It used a standard language model objective, not designed for cross-lingual purposes. Even though, it achieved surprising results which inspired many scientists to investigate this direction. \newcite{LASER} published a multilingual language model (henceforth LASER) which outperformed m-BERT \cite{mBERT}. The model was trained on as many as 93 languages using a machine translation objective with BiLSTMs. Another interesting model \cite{Multifit}(henceforth MultiFit) is monolingual, but thanks to a bootstrapping method \cite{Ruder2018}, it works well with zero-shot learning. Namely, it employs LASER to predict so called pseudo-labels of the dataset in the target language and by using them in training it actually surpasses LASER performance. 

The most recent model was proposed by \newcite{xlmr}, dubbed XLM-R. It is based on the RoBERTa architecture \cite{roberta}. XLM-R has a translation language modeling (TLM) objective \cite{xlmr} in which parallel sentences are fed together as examples and by masking some words in one language the model indirectly uses the translation from the other language. It is the first multilingual model that matches performance of strong monolingual models \cite{xlmr}.
For our experiments we used XLM-R with a newly published cross-lingual benchmark called XTREME \cite{xtreme} which allows to easily evaluate XLM-R on up to 40 languages and 8 diverse NLP tasks.

\subsection{What determines good transfer performance?}
\newcite{Lin2019} created an automatic approach to predict transfer performance in zero-shot learning setups. They examined the impact of a wide variety of factors, ranging from  word overlap or type-token ratio in the training set to geographic, genetic or syntactic distance between languages. They evaluated their approach on four tasks: machine translation, entity linking, part of speech tagging and dependency parsing and for each they used a task-specific architecture without incorporating any pretrained embeddings. 
Two other papers performed various experiments connected to a similar topic on m-BERT \cite{mBERT} using this single model on different tasks. \newcite{Pires2019} focused on named entity recognition (NER) and part-of-speech tagging (POS) while \newcite{Tran2019} covered dependency parsing. Below we discuss their findings grouped by different factors.

\subsubsection{Word overlap}

Results of \newcite{Lin2019} show that word overlap seems to be one of the best features which suggests that it is related to transfer - high word overlap between the source and the target language leads to good performance. 
On the other hand, \newcite{Pires2019} showed that there is little correlation between word overlap and performance of m-BERT (however this relation was strongly linear for the original version of BERT that was only trained on English). They used NER because they expected "zero-shot performance on NER to be highly dependent on word piece overlap"\cite{Pires2019}. These results are confirmed by \newcite{Tran2019} on dependency parsing.

\subsubsection{Dataset size}
\newcite{Lin2019} uses as one of the features the ratio between training size of the source and training languages. It seems to have very little correlation with the performance for all evaluated tasks. In our view, the size of the training set is a much more important factor than the ratio. Let us consider if training sets of both source and target languages are big and of similar size, then the ratio is large and the performance will probably be high. If we take two small languages, then the ratio is also large, but the performance will probably be low. This clearly suggests that this measure will not be strongly correlated with the performance, without doing any experiments.
\newcite{xtreme} calculated Pearson correlation between Wikipedia size (which is not used for pretraining XLM-R) and performance on all their tasks and the outcome is very interesting. For both text classification datasets (XNLI \cite{xnli} and PAWSX \cite{pawsx}) the correlation was very strong - about 0.80, while for POS \cite{Nivre2018}
and NER \cite{pan2017} it was as little as 0.30. This indicates that the significance of dataset size varies significantly between tasks.

\subsubsection{Syntactic features}
There is a lot of evidence that syntax helps in NLP tasks such as POS tagging or dependency parsing (\citealp{ammar2016}; \citealp{naseem2012}; \citealp{zhang2015}; \citealp{ostling2015}). However, it is usually explicitly incorporated as features which is not the case for models analyzed by (\citealp{Lin2019}; \citealp{Pires2019}; \citealp{Tran2019}). 

WALS (World Atlas of Language Structures) \cite{wals} is a very useful database containing 192 linguistic features for 2662 different languages which was often used in previous literature. \newcite{Pires2019} grouped languages with regard to 6 WALS features regarding word order and concluded that the more features the source and target languages have in common, the better the performance. \newcite{Tran2019} showed that in their experiments syntactic distance is not correlated with performance. Syntactic distance is computed as follows: for each language in the source-target language pair a one-hot encoded vector with WALS syntactic features is retrieved and then a distance between these two vectors is computed. 
In experiments performed by \newcite{Lin2019}, syntactic distance was the 4th (out of 10) most important feature after word overlap, genetic and geographical distance respectively. 

When using syntactic distance, all WALS features contribute equally to the score, however in reality particular features have different importance. Previously, most often only the features related to word order were used \cite{survey} which suggests that these features are more important than others and shouldn't be treated with equal weights. That is why in this work we examine separate features and how using them compares to an aggregated distance in predicting the transfer quality. Moreover, to verify if the significance of different factors varies among downstream task we cover POS tagging, named entity recognition and text classification.
Furthermore, as opposed to \newcite{Lin2019}, we use a single, state-of-the-art language model (XLM-R) for all downstream tasks which decreases the chance that the differences in performance are due to differences in model architecture rather than tasks.

\section{Datasets}

\subsection{Universal Dependencies 2.0 (POS tagging)}
We use Universal Dependencies \cite{Nivre2018} for part of speech tagging on 29 languages. There are 17 tags (categories) universal for all languages.\footnote{To learn more about UD tags and what each abbreviation means visit: https://universaldependencies.org/u/pos/}

\subsection{Panx (NER)}
For named entity recognition, we use Wikiann \cite{pan2017}, more  precisely the balanced sample made by \cite{rahimi-etal-2019-massively}. There are three types of entities: Location, Person and Organization. The entities were automatically labeled from Wikipedia.

\subsection{XNLI (NLI)}
This task \cite{xnli} concerns natural language inference (NLI), i.e. for each example there is a pair of sentence and there are three possible labels describing the relationship of this pair: contradiction, entailment or neutral. It is a special case of text classification, yet it is substantially more difficult than standard tasks from this group. For instance, while for sentiment analysis often one strongly polarized word is enough to predict the correct class, here the distinction between two sentences in a pair is much more subtle. The dataset is available in 15 languages. Due to its large size and limited resources we downsample randomly each training set from almost 400k examples to 100k (since the classes are balanced in the original dataset, after random sampling they remain balanced).

\begin{table*}[t]
\small
\centering
\begin{tabular}{l|l|l|l|l|l|l|l|l|l|l|l|l|l|l|l}
\textbf{test } &    \# & \textbf{train } &    \textbf{acc } & \textbf{test } &    \# & \textbf{train } &    \textbf{acc } & \textbf{test } &    \# & \textbf{train } &    \textbf{acc } & \textbf{test } &    \# & \textbf{train } &    \textbf{acc } \\
\hline
  af &  sup &    af &  0.984 &   et &  sup &    et &  0.979 &   id &  sup &    id &  0.936 &   ru &  sup &    ru &  0.985 \\
  af &    1 &    en &  0.884 &   et &    1 &    fi &  0.905 &   id &    1 &    en &  0.856 &   ru &    1 &    bg &  0.912 \\
  af &    2 &    nl &  0.876 &   et &    2 &    en &  0.894 &   id &    2 &    pt &  0.853 &   ru &    2 &    it &  0.906 \\
  af &    3 &    de &  0.865 &   et &    3 &    es &  0.881 &   id &    3 &    it &  0.847 &   ru &    3 &    fr &  0.905 \\
\hline
  ar &  sup &    ar &  0.966 &   eu &  sup &    eu &  0.964 &   it &  sup &    it &  0.976 &   ta &  sup &    ta &  0.849 \\
  ar &    1 &    he &  0.744 &   eu &    1 &    ur &  0.788 &   it &    1 &    fr &  0.935 &   ta &    1 &    et &  0.800 \\
  ar &    2 &    fa &  0.729 &   eu &    2 &    hi &  0.784 &   it &    2 &    es &  0.922 &   ta &    2 &    de &  0.762 \\
  ar &    3 &    ru &  0.702 &   eu &    3 &    nl &  0.775 &   it &    3 &    pt &  0.910 &   ta &    3 &    fi &  0.759 \\
\hline
  bg &  sup &    bg &  0.991 &   fa &  sup &    fa &  0.985 &   ja &  sup &    ja &  0.980 &   te &  sup &    te &  0.941 \\
  bg &    1 &    fr &  0.910 &   fa &    1 &    ru &  0.778 &   ja &    1 &    zh &  0.553 &   te &    1 &    en &  0.890 \\
  bg &    2 &    it &  0.906 &   fa &    2 &    it &  0.776 &   ja &    2 &    ko &  0.508 &   te &    2 &    hi &  0.873 \\
  bg &    3 &    ru &  0.904 &   fa &    3 &    ar &  0.771 &   ja &    3 &    nl &  0.432 &   te &    3 &    et &  0.869 \\
\hline
  de &  sup &    de &  0.980 &   fi &  sup &    fi &  0.957 &   ko &  sup &    ko &  0.941 &   tr &  sup &    tr &  0.960 \\
  de &    1 &    nl &  0.897 &   fi &    1 &    et &  0.890 &   ko &    1 &    eu &  0.630 &   tr &    1 &    hi &  0.805 \\
  de &    2 &    en &  0.885 &   fi &    2 &    en &  0.872 &   ko &    2 &    te &  0.628 &   tr &    2 &    ur &  0.803 \\
  de &    3 &    it &  0.846 &   fi &    3 &    it &  0.861 &   ko &    3 &    en &  0.625 &   tr &    3 &    eu &  0.797 \\
\hline
  el &  sup &    el &  0.979 &   he &  sup &    he &  0.972 &   mr &  sup &    mr &  0.840 &   ur &  sup &    ur &  0.949 \\
  el &    1 &    de &  0.877 &   he &    1 &    fi &  0.788 &   mr &    1 &    de &  0.860 &   ur &    1 &    hi &  0.910 \\
  el &    2 &    en &  0.870 &   he &    2 &    et &  0.785 &   mr &    2 &    ur &  0.852 &   ur &    2 &    hu &  0.743 \\
  el &    3 &    ru &  0.867 &   he &    3 &    it &  0.776 &   mr &    3 &    et &  0.848 &   ur &    3 &    it &  0.740 \\
\hline
  en &  sup &    en &  0.966 &   hi &  sup &    hi &  0.981 &   nl &  sup &    nl &  0.979 &   vi &  sup &    vi &  0.925 \\
  en &    1 &    it &  0.881 &   hi &    1 &    ur &  0.944 &   nl &    1 &    en &  0.914 &   vi &    1 &    fr &  0.633 \\
  en &    2 &    es &  0.871 &   hi &    2 &    tr &  0.788 &   nl &    2 &    fr &  0.904 &   vi &    2 &    id &  0.629 \\
  en &    3 &    nl &  0.870 &   hi &    3 &    he &  0.774 &   nl &    3 &    it &  0.904 &   vi &    3 &    he &  0.627 \\
\hline
  es &  sup &    es &  0.977 &   hu &  sup &    hu &  0.971 &   pt &  sup &    pt &  0.978 &   zh &  sup &    zh &  0.966 \\
  es &    1 &    it &  0.943 &   hu &    1 &    nl &  0.863 &   pt &    1 &    es &  0.935 &   zh &    1 &    ja &  0.662 \\
  es &    2 &    fr &  0.938 &   hu &    2 &    de &  0.862 &   pt &    2 &    it &  0.926 &   zh &    2 &    he &  0.574 \\
  es &    3 &    pt &  0.929 &   hu &    3 &    es &  0.862 &   pt &    3 &    fr &  0.920 &   zh &    3 &    ko &  0.569 \\

\end{tabular}
\caption{Top 3 languages (train) for each target language (test) together with the supervised result on POS tagging (accuracy)}
\label{fig:udpos_results}
\end{table*}

\section{Analysis}

\begin{table*}[t]
\small
\centering
\input{ner_results}
\caption{Top 3 languages (train) for each target language (test) together with its supervised result on NER (accuracy)}
\label{fig:ner_results}
\end{table*}
\begin{table*}[t]
\small
\centering
\input{xnli_results}
\caption{Top 3 languages (train) for each target language (test) together with its supervised result on NLI (accuracy)}
\label{fig:xnli_results}
\end{table*}

We use the XTREME framework to evaluate 3 standard NLP downstream tasks on XLM-R for all available languages (and trying all source-target pairs possible). As we used 28 languages for POS tagging, 37 languages for NER and 15 for XNLI showing results for all pairs wouldn't be feasible, therefore in the paper we decided to show for each test language its supervised result compared with top 3 source languages in Table \ref{fig:udpos_results}, Table \ref{fig:ner_results} and Table \ref{fig:xnli_results}. For abbreviation, we use ISO 693-1 language codes\footnote{Check https://en.wikipedia.org/wiki/List\_of\_ISO\_639\-1\_codes to see how codes correspond to languages}.

\subsection{Results on downstream tasks}
\label{downstream_results}
For POS tagging (Table \ref{fig:udpos_results}), some outcomes are in line with the expectations (e.g. Portuguese (pt) with the top-3 languages as Spanish (es), Italian (it) and French (fr)) while others are not. All languages apart from Marathi (mr) obtain better results in supervised setting than zero-shot. Its best source language is German (de) which is quite surprising as Marathi is part of the Indic language family represented in the dataset by two other languages - Urdu (ur) and Hindi (hi). French and Italian obtain better results on Bulgarian (a Slavic language) than the other Slavic language in the dataset, Russian. Other interesting pairs include German being the best language for Greek (el), Urdu the best for Basque (eu) - although not that surprising keeping in mind that Basque is a language isolate, Dutch (nl) the best for Hungarian (another language isolate), Basque the best for Korean (ko) and Estonian (et) best for Tamil (ta, a Dravidian language spoken in India and Sri Lanka). Interestingly, some of the results seem related to colonialism - for instance English being the best language for Indonesian (id) and French being the best for Vietnamese (vi).

While for POS tagging many outcomes were clear and we only pointed out the outliers, it is much more difficult to find patterns for NER (Table \ref{fig:ner_results}). Hungarian is the third best for Afrikaan and German, this time Italian is the best for Greek, Indonesian for English, French for Farsi (fa) and Italian (instead of a Germanic langauge) for Dutch. 

For XNLI (Table \ref{fig:xnli_results}) as many as 5 out of 15 languages obtain worse results in supervised setting than zero-shot: Bulgarian, Spanish, Hindi, Russian and Urdu. We can also notice that XNLI is more challenging than POS or NER - while for the first two tasks accuracy is generally above 0.80, for XNLI, none of the language pairs exceeds that number.

After some initial, qualitative remarks, we follow up with more quantitative analysis.
To investigate whether the downstream task is important in choosing the source language we inspect Table \ref{fig:correlations_tasks} which shows that accuracy between NER and POS tasks is much stronger than between those two and XNLI. This means that there is no guarantee that if a source language achieves good performance on a target language for one NLP task, it will also perform well on a different task.

Table \ref{fig:correlations_dist} presents how performance on downstream tasks correlates with three most important linguistic distance measures based on \cite{Lin2019}: syntactic (syn), geographic (geo) and genetic (gen) distances. As mentioned in previous sections, syntactic distance is the aggregate of a group of linguistic features, geographic distance is the distance between geographic locations where two languages are natively spoken while genetic distance expresses how related the languages are - French and Spanish are closely related as they both belong to the Romance language family while French and Chinese are from different language families, thus their genetic distance is larger. We can clearly see that POS is strongly correlated with all three distances. XNLI correlates the most with geographical distance and for NER all correlations are relatively weak. In the following section we focus on particular linguistic features and examine their importance to explore their effect on transfer between languages.

\begin{table}
\renewcommand{\arraystretch}{1.1}%
\begin{tabular}{l|l|l|l}
  task &       syn &       geo &       gen \\
\hline

NER & -0.183454 & -0.197949 & -0.198537 \\
POS & -0.292522 & -0.378323 & -0.319140 \\
NLI & -0.157961 & -0.274467 & -0.100264 \\
\end{tabular}
\caption{Performance correlation between tasks and distances}
\label{fig:correlations_dist}
\end{table}%

\begin{table}

\renewcommand{\arraystretch}{1.1}%
\begin{tabular}{l|l|l|l}

{}  &       NER &       POS &      NLI \\
\hline
NER &  1.000000 &  0.495913 &  0.179361 \\
POS &  0.495913 &  1.000000 &  0.065942 \\
NLI &  0.179361 &  0.065942 &  1.000000 \\

\end{tabular}
\caption{Performance correlation between tasks}\label{tab1}
\label{fig:correlations_tasks}
\end{table}%

\subsection{Feature importance for transfer learning}
To measure the importance of particular linguistic features (more specifically WALS features) with regard to performance we use two approaches. The first approach uses a statistical test. ANOVA is often used to measure correlation between a categorical independent variable and a continuous dependent variable in many fields, including NLP (\citealp{saha-das-mandal-2015-analysis}; \citealp{li-etal-2014-assessing}; \citealp{malmasi-dras-2014-visualisation}). However, we cannot make assumptions about distribution of our data, so we use Kruskal-Wallis test which is a non-parametric equivalent of ANOVA and tests if the difference in means between groups is significant. %The more important is the feature, the smaller is the p-value of its test.
Since we are mostly interested in choosing the right source language for a given target language, we examined the correlation between features of the source language and the accuracy. The second approach measures feature importance after training a simple, interpretable machine learning (ML) model.  It is very important to distinguish the models used in the Section \ref{downstream_results} and in this section. In Section \ref{downstream_results} we used a multilingual language model, XLM-R, to make predictions on 3 NLP downstream tasks: POS tagging, NER and NLI. In this case, we use an interpretable model such as Random Forest on tabular data: A single data point fed into this model consists of the source language, the target language, linguistic features of both languages and the score for one of the downstream tasks predicted by the first model. Therefore, the second model aims to predict the performance of the first model using linguistic features. Although we focus on feature analysis, as a side effect we can use the second model to predict transfer between languages. In Section \ref{comparison} we show that using such a model allows better prediction of transfer performance than only using syntactic distance.

Both approaches have pros and cons. The former method only captures simple correlation patterns. The latter method can capture non-linear correlations, but its reliability depends on the quality of the model. Results for both methods are presented in the following sections.

\subsubsection{Feature importance with Kruskal-Wallis test}
% , the most significant features are 90A (Order of Relative Clause and Noun) and 96A (Relationship between the Order of Object and Verb and the Order of Relative Clause and Noun). 

We can see the important features indicated by Kruskal-Wallis test in Table \ref{fig:kw_test}. The most important features for POS taging align with the results obtained with the ML method when using one-hot encoding (discussed in the next subsection). The first feature for NER is also found to be important by the second method. Similarly, XNLI has one overlapping feature between two methods. 

% Moreover, we find the best features for each language separately and discover interesting patterns. Namely, for all Romance languages feature 86A (Order of Genitive and Noun) was the most important while for all Slavic and Uralic languages the most signifcant was 93A (Position of Interrogative Phrases in Content Questions).

% When we look for best features per language, both 121A (Comparative Constructions) and 138A (Tea) appear to be the best features for 5 languages. As opposed to POS tagging, it is not trivial to find relations between languages for which the same feature is best.

% When considering all languages together, the most significant features are  90A (Order of Relative Clause and Noun) and 45A (Politeness Distinctions in Pronouns). When examined per language, almost all languages have different important features.
\begin{table}[t]
\input{kruskal_wallis}
\caption{Top-3 important features according to Kruskal-Wallis test per downstream task. Features with * overlap with important features found with the ML method.}
\label{fig:kw_test}
\end{table}

\subsubsection{Feature importance with ML models}
To evaluate feature importance we create a dataset where each row is corresponding to a train-test language pair together with its score and linguistic features of both languages. We train different interpretable models (Linear Regression, Random Forest and XGBoost) that predict accuracy of a given setting and as a result obtain feature importance values that show which language features determine the score in the most accurate way. We deleted rows for which the training and test language is the same because we are focusing on zero-shot learning. From all available WALS features \cite{wals} we only discarded phonological features as they are irrelevant for text processing together with features that had missing values for all languages \footnote{Due to some problems with the data, we also remove any pairs containing Japanese or Chinese, pairs when French is the target language and the German - English pair}. 
As aforementioned, such a model not only allows us to interpret linguistic features, but also can be used to predict the performance for given source and target languages. Table \ref{fig:2nd_model_errors} renders the results. For POS tagging the obtained test RMSE is 0.0423 which can be roughly interpreted that on average our model was off by about 4\% of accuracy. For NER it is quite similar - 0.0468. For XNLI, which has a smaller subset of languages and lower variance between scores we achieved RMSE of 0.007936 (with linear regression). 
We experimented with two approaches for encoding WALS features - ordinal encoding - where we assign an integer to each category and one-hot encoding. Both achieved similar RMSE, however feature importance varies between those two so below we selected features that were at the top for both approaches.
Table \ref{fig:feat_importance} indicates the top-5 important features per task and encoding. Features that were found as important by both encodings or by both Kruskal-Wallist test and one of the encodings are presented in Table \ref{fig:feat_explained}. For each feature we present the significant category that tends to yield lower accuracy and languages that fall in that category. This allows us to suggest rules-of-thumb for predicting the performance, for instance: "If the source language has Object-Verb and Relative Clause-Noun orders, it might lead to low accuracy, therefore one should avoid using languages in this group".

%For UD POS and NER the best performance was obtained by a random forest, while for XNLI - linear regression. This is probably the case because interpretable methods are more prone to overfitting than linear regression and XNLI covers less languages and therefore has less training examples.

\begin{table}[t]
\begin{tabular}{l|l|l}
Task  & Linguistic features &  Syntactic distance \\
\hline
POS & 0.0423 & 0.10 \\
NER & 0.0468 & 0.12 \\
XNLI & 0.007936 & 0.04 \\
\end{tabular}
\caption{Comparison of models predicting transfer performance for all available language pairs (RMSE)}
\label{fig:2nd_model_errors}
\end{table}

\begin{table}[t]
\small
\centering
\begin{tabular}{l|l|l|l|l}
     Task &     Encoding   &  \# &                                           Feature &        Importance \\
% task &  &    &                                                   &                   \\

\hline
POS & ordinal &  1 &                                          55A\_test &           0.01518 \\
     &  &  2 &                                         94A\_train &           0.01364 \\
     &  &  3 &                                         36A\_train &           0.00551 \\
     &  &  4 &                                         138A\_test &           0.00489 \\
     &  &  5 &                                         93A\_train &           0.00324 \\
     & onehot &  1 &  90B\_train\_1 &           0.00611 \\
     &  &  2 &                           96A\_train\_1 &           0.00610 \\
     &  &  3 &                  90A\_train\_2 &           0.00606 \\
     &  &  4 &      93A\_train\_2 &           0.00542 \\
     &  &  5 &                               138A\_test\_0 &           0.00537 \\
     \hline
NER & ordinal &  1 &                                          89A\_test &           0.02715 \\
     &  &  2 &                                          66A\_test &           0.02006 \\
     &  &  3 &                                         20A\_train &           0.00795 \\
     & onehot &  1 &                           89A\_test\_2 &           0.02702 \\
     &  &  2 &                          66A\_test\_4 &           0.01039 \\
     &  &  3 &                       118A\_test\_1 &           0.00870 \\
     \hline
XNLI & ordinal &  1 &                                         45A\_train &  5.69e+09 \\
     &  &  2 &                                         44A\_train &  3.94e+09 \\
     &  &  3 &                                         27A\_train &  3.26e+09 \\
     & onehot &  1 &               22A\_train\_3 &  1.53e+10 \\
     &  &  2 &                               42A\_train\_0 &  1.02e+10 \\
     &  &  3 &                     25A\_train\_2 &  9.85e+09 \\
\end{tabular}
\caption{Most important features according to ML models per downstream task.}
\label{fig:feat_importance}
\end{table}

\begin{table*}[t]
\small
\centering
\begin{tabular}{p{1cm}|p{5cm}|p{4cm}|p{5cm}}
     Task & Feature & Significant category & Languages  \\

\hline
POS & 96A Relationship between the Order of Object and Verb and the Order of Relative Clause and Noun &  source language with Object-Verb and  Relative  Clause-Noun  orders & Basque, Korean, Malayalam,Marathi, Burmese, Tamil, Telugu, Turkish \\
\hline
& 93A  Interrogative  Phrases  in  Content  Questions
%\footnote{Examples of interrogative phrases in English are: where, what, who etc.} 
& source language with not initial Interrogative Phrase & Basque, Persian,  Hindi,  Hungarian,  Japanese, Georgian,Korean, Malayalam, Marathi, Burmese, Swahili,Tamil, Telugu, Thai, Turkish, Urdu, Mandarin \\
\hline
NER & 89A: Order of Numeral and Noun & target language with Noun-Numeral order & Burmese, Swahili, Thai, Yoruba \\
\hline
& 66A The past tense 
% \footnote{Most European languageshave special grammatical constructions that explic-itly indicate if a sentence is in present or past tense.However, there is a large group of languages thatdo not have such distinction and the tense is de-rived from context} 
& source language with no present-past tense distinction & Indonesian, Burmese, Thai, Tagalog, Vietnamese, Yoruba \\
\hline
XNLI & 45A: Politeness Distinctions in Pronouns & source language with multiple politeness distinctions or pronouns avoided for politeness & Hindi, Urdu, Thai, Vietnamese

\end{tabular}
\caption{Selected important features with categories that yield lower performance.}
\label{fig:feat_explained}
\end{table*}

\subsubsection{Feature importance by feature group}
WALS features are divided in multiple groups: Morphology, Nominal Categories, Nominal Syntax, Verbal Categories, Word Order, Simple Clauses, Complex Sentences, Lexicon, Sign Languages and Others. We separately trained our best model on each group to determine which one is the most salient. For POS tagging and NER, the winning category is Word Order (with RMSE of 0.0401 and 0.0487 respectively) which confirms previous literature. For XNLI, Word order is much less significant (only the 4th best result among groups) and the best group is Verbal Categories (RMSE of 0.00772).

\subsubsection{Comparing fine-grained linguistic features to URIEL syntactic distance}
\label{comparison}

Furthermore, we move to answer the question whether using fine-grained linguistic features helps more in predicting the zero-shot performance compared to an aggregated syntactic distance used by \cite{Lin2019}. To verify that we use the same syntactic distance computed with the lang2vec library \cite{uriel}. As shown in Table \ref{fig:2nd_model_errors}, in all experiments syntactic distance itself was worse at predicting the zero-shot performance than the separate features (with RMSE equal to 0.10, 0.12 and 0.04 for POS tagging, NER and XNLI respectively). Therefore we conclude that replacing a syntactic distance with fine-grained features in systems such as LangRank \cite{Lin2019} would most certainly improve predictions.

\section{Conclusion}
In this paper we explore the effect of linguistic features (for different languages) on zero-short learning with the goal to identify which language is best to transfer from. 

Importance of particular features appears to heavily depend on the downstream task - there is no single WALS feature that would be the most important for multiple tasks. This insight followed by the computed correlations between tasks shows that one  should not expect that for  a  target language $L1$ there is a single language $L2$ that is the best choice for any NLP  task. 

Most importantly, we prove that using fine-grained linguistic features allows to predict the performance 2-4 times better than an aggregated syntactic distance and confirm that word order is paramount for POS tagging. Therefore in automatic systems like the one proposed by \newcite{Lin2019} particular linguistic features should be included to improve performance. 
% As oppose to Lin et al, 2019, we don't use any dataset-specific features (which cannot be used when deciding for which language the dataset will be collected such as type/token ratio or word overlap) which makes our work more universal.

Although this work sheds some light on the unexplored topic of the effect of linguistic features on transfer learning, there is still much work to be done. One of the issues is that apart from POS tagging, NER and dependency parsing there are hardly any tasks with datasets covering more than 15 languages. Using more languages would make the results more reliable. To encourage the community to perform further analyses we publish the predictions and the code that was used for analysis. Moreover, we acknowledge that there might be other, more complex factors than the ones discussed in this paper that provide better explanation to the performance differences between languages. Therefore, as this topic lays in the intersection of natural language processing and linguistics, cooperation between experts from both fields can lead to interesting findings.
\\

\end{document}

%% file: ner_results.tex
\begin{tabular}{l|l|l|l|l|l|l|l|l|l|l|l|l|l|l|l}
\textbf{test } &    \# & \textbf{train } &    \textbf{acc } & \textbf{test } &    \# & \textbf{train } &    \textbf{acc } & \textbf{test } &    \# & \textbf{train } &    \textbf{acc } & \textbf{test } &    \# & \textbf{train } &    \textbf{acc } \\
\hline

  af &  sup &    af &  0.978 &   fa &  sup &    fa &  0.971 &   kk &  sup &    kk &  0.957 &   ta &  sup &    ta &  0.958 \\
  af &    1 &    nl &  0.953 &   fa &    1 &    fr &  0.807 &   kk &    1 &    af &  0.912 &   ta &    1 &    ka &  0.871 \\
  af &    2 &    it &  0.942 &   fa &    2 &    hi &  0.794 &   kk &    2 &    id &  0.899 &   ta &    2 &    mr &  0.868 \\
  af &    3 &    hu &  0.941 &   fa &    3 &    ur &  0.778 &   kk &    3 &    fi &  0.889 &   ta &    3 &    ko &  0.867 \\
\hline
  ar &  sup &    ar &  0.957 &   fi &  sup &    fi &  0.971 &   ko &  sup &    ko &  0.960 &   te &  sup &    te &  0.938 \\
  ar &    1 &    fa &  0.826 &   fi &    1 &    hu &  0.940 &   ko &    1 &    el &  0.862 &   te &    1 &    ta &  0.909 \\
  ar &    2 &    hi &  0.810 &   fi &    2 &    et &  0.936 &   ko &    2 &    ka &  0.858 &   te &    2 &    ml &  0.896 \\
  ar &    3 &    de &  0.807 &   fi &    3 &    nl &  0.934 &   ko &    3 &    mr &  0.856 &   te &    3 &    hu &  0.888 \\
\hline
  bg &  sup &    bg &  0.976 &   fr &  sup &    fr &  0.951 &   ml &  sup &    ml &  0.952 &   th &  sup &    th &  0.914 \\
  bg &    1 &    ru &  0.933 &   fr &    1 &    pt &  0.908 &   ml &    1 &    ta &  0.909 &   th &    1 &    zh &  0.738 \\
  bg &    2 &    hu &  0.926 &   fr &    2 &    it &  0.907 &   ml &    2 &    ka &  0.895 &   th &    2 &    ja &  0.721 \\
  bg &    3 &    nl &  0.924 &   fr &    3 &    es &  0.902 &   ml &    3 &    el &  0.892 &   th &    3 &    my &  0.715 \\
\hline
  bn &  sup &    bn &  0.978 &   he &  sup &    he &  0.957 &   mr &  sup &    mr &  0.960 &   tl &  sup &    tl &  0.972 \\
  bn &    1 &    ta &  0.820 &   he &    1 &    ar &  0.863 &   mr &    1 &    ka &  0.896 &   tl &    1 &    vi &  0.808 \\
  bn &    2 &    hi &  0.797 &   he &    2 &    it &  0.856 &   mr &    2 &    hi &  0.892 &   tl &    2 &    en &   0.79 \\
  bn &    3 &    ml &  0.793 &   he &    3 &    el &  0.845 &   mr &    3 &    ar &  0.885 &   tl &    3 &    fr &  0.775 \\
\hline
  de &  sup &    de &  0.970 &   hi &  sup &    hi &  0.945 &   ms &  sup &    ms &  0.981 &   ur &  sup &    ur &  0.988 \\
  de &    1 &    nl &  0.939 &   hi &    1 &    mr &  0.866 &   ms &    1 &    vi &  0.857 &   ur &    1 &    ar &  0.852 \\
  de &    2 &    it &  0.927 &   hi &    2 &    ka &  0.864 &   ms &    2 &    hu &  0.844 &   ur &    2 &    fa &  0.845 \\
  de &    3 &    hu &  0.926 &   hi &    3 &    ta &  0.855 &   ms &    3 &    it &  0.841 &   ur &    3 &    ml &  0.843 \\
\hline
  el &  sup &    el &  0.975 &   hu &  sup &    hu &  0.977 &   my &  sup &    my &  0.740 &   vi &  sup &    vi &  0.959 \\
  el &    1 &    it &  0.931 &   hu &    1 &    fi &  0.949 &   my &    1 &    ka &  0.866 &   vi &    1 &    id &  0.857 \\
  el &    2 &    nl &  0.930 &   hu &    2 &    nl &  0.943 &   my &    2 &    ko &  0.865 &   vi &    2 &    fr &  0.841 \\
  el &    3 &    fi &  0.929 &   hu &    3 &    it &  0.942 &   my &    3 &    ta &  0.860 &   vi &    3 &    nl &  0.832 \\
\hline
  en &  sup &    en &  0.929 &   id &  sup &    id &  0.969 &   nl &  sup &    nl &  0.971 &   zh &  sup &    zh &  0.933 \\
  en &    1 &    id &  0.863 &   id &    1 &    vi &  0.894 &   nl &    1 &    it &  0.935 &   zh &    1 &    ja &  0.883 \\
  en &    2 &    vi &  0.861 &   id &    2 &    ms &  0.884 &   nl &    2 &    hu &  0.931 &   zh &    2 &    te &  0.811 \\
  en &    3 &    nl &  0.858 &   id &    3 &    et &  0.863 &   nl &    3 &    fi &  0.925 &   zh &    3 &    ta &  0.807 \\
\hline
  es &  sup &    es &  0.950 &   it &  sup &    it &  0.966 &   pt &  sup &    pt &  0.954 &      &      &       &        \\
  es &    1 &    fr &  0.921 &   it &    1 &    nl &  0.924 &   pt &    1 &    es &  0.918 &      &      &       &        \\
  es &    2 &    pt &  0.920 &   it &    2 &    et &  0.915 &   pt &    2 &    fr &  0.902 &      &      &       &        \\
  es &    3 &    it &  0.910 &   it &    3 &    fi &  0.914 &   pt &    3 &    it &  0.902 &      &      &       &        \\
\hline
  et &  sup &    et &  0.976 &   ja &  sup &    ja &  0.900 &   ru &  sup &    ru &  0.960 &      &      &       &        \\
  et &    1 &    fi &  0.943 &   ja &    1 &    zh &  0.828 &   ru &    1 &    bg &  0.889 &      &      &       &        \\
  et &    2 &    hu &  0.929 &   ja &    2 &    th &  0.773 &   ru &    2 &    et &  0.885 &      &      &       &        \\
  et &    3 &    nl &  0.924 &   ja &    3 &    hi &  0.760 &   ru &    3 &    nl &  0.873 &      &      &       &        \\
\hline
  eu &  sup &    eu &  0.974 &   ka &  sup &    ka &  0.965 &   sw &  sup &    sw &  0.940 &      &      &       &        \\
  eu &    1 &    hu &  0.922 &   ka &    1 &    el &  0.921 &   sw &    1 &    tl &  0.859 &      &      &       &        \\
  eu &    2 &    af &  0.915 &   ka &    2 &    fi &  0.918 &   sw &    2 &    vi &  0.823 &      &      &       &        \\
  eu &    3 &    et &  0.914 &   ka &    3 &    hu &  0.918 &   sw &    3 &    id &  0.815 &      &      &       &        \\

\end{tabular}

%% file: xnli_results.tex
\begin{tabular}{l|l|l|l|l|l|l|l|l|l|l|l|l|l|l|l}

\textbf{test } &    \# & \textbf{train } &    \textbf{acc } & \textbf{test } &    \# & \textbf{train } &    \textbf{acc } & \textbf{test } &    \# & \textbf{train } &    \textbf{acc } & \textbf{test } &    \# & \textbf{train } &    \textbf{acc } \\
\hline

  ar &  sup &    ar &  0.718 &   en &  sup &    en &  0.815 &   ru &  sup &    ru &  0.739 &   ur &  sup &    ur &  0.624 \\
  ar &    1 &    bg &  0.713 &   en &    1 &    de &  0.813 &   ru &    1 &    vi &  0.752 &   ur &    1 &    zh &  0.683 \\
  ar &    2 &    el &  0.713 &   en &    2 &    fr &  0.806 &   ru &    2 &    bg &  0.738 &   ur &    2 &    ru &  0.677 \\
  ar &    3 &    ru &  0.711 &   en &    3 &    bg &  0.804 &   ru &    3 &    es &  0.736 &   ur &    3 &    th &  0.673 \\
\hline
  bg &  sup &    bg &  0.747 &   es &  sup &    es &  0.763 &   sw &  sup &    sw &  0.656 &   vi &  sup &    vi &   0.74 \\
  bg &    1 &    th &  0.759 &   es &    1 &    ru &  0.780 &   sw &    1 &    de &  0.645 &   vi &    1 &    bg &  0.739 \\
  bg &    2 &    ru &  0.751 &   es &    2 &    de &  0.777 &   sw &    2 &    el &  0.643 &   vi &    2 &    de &  0.738 \\
  bg &    3 &    vi &  0.748 &   es &    3 &    zh &  0.768 &   sw &    3 &    th &  0.636 &   vi &    3 &    es &  0.737 \\
\hline
  de &  sup &    de &  0.763 &   fr &  sup &    fr &  0.761 &   th &  sup &    th &  0.746 &   zh &  sup &    zh &  0.749 \\
  de &    1 &    fr &  0.757 &   fr &    1 &    el &  0.760 &   th &    1 &    zh &  0.730 &   zh &    1 &    th &  0.748 \\
  de &    2 &    ar &  0.755 &   fr &    2 &    de &  0.759 &   th &    2 &    bg &  0.728 &   zh &    2 &    ru &  0.744 \\
  de &    3 &    tr &  0.755 &   fr &    3 &    ru &  0.759 &   th &    3 &    ar &  0.727 &   zh &    3 &    de &  0.739 \\
\hline
  el &  sup &    el &  0.748 &   hi &  sup &    hi &  0.701 &   tr &  sup &    tr &  0.735 &      &      &       &        \\
  el &    1 &    de &  0.747 &   hi &    1 &    ru &  0.707 &   tr &    1 &    th &  0.725 &      &      &       &        \\
  el &    2 &    es &  0.745 &   hi &    2 &    th &  0.705 &   tr &    2 &    fr &  0.724 &      &      &       &        \\
  el &    3 &    fr &  0.745 &   hi &    3 &    tr &  0.705 &   tr &    3 &    bg &  0.722 &      &      &       &        \\

\end{tabular}

%% file: kruskal_wallis.tex
\begin{tabular}{l|l|l|l|l}

  Task & Feature &  \# & h & p-value \\

\hline

 POS &  90A\_train* &  1 &  162.88 &  4.39e-35 \\
  &  96A\_train* &  2 &  162.29 &  4.73e-34 \\
  &  53A\_train &  3 &  146.53 &  1.13e-30 \\
 \hline
   NER &  89A\_train* &  1 &   65.40 &  6.11e-16 \\
    &  60A\_train &  2 &   76.35 &  1.03e-15 \\
    &  20A\_train &  3 &   64.64 &  5.00e-14 \\
   \hline
  XNLI &  90A\_train &  1 &   14.93 &  5.73e-04 \\
   &  45A\_train* &  2 &   17.22 &  6.37e-04 \\
   &  49A\_train &  3 &   15.54 &  3.71e-03 \\

\end{tabular}